# Synthèse non quadratique H∞ de contrôleurs décentralisés pour un ensemble de descripteurs flous T-S interconnectés

# H∞ decentralized non quadratic controller design for a set of interconnected Takagi-Sugeno descriptors

D. Jabri[1,2], K. Guelton[1], N. Manamanni[1], M.N. Abdelkrim[2]

[1] Centre de Recherche en Sciences et Technologies de l'Information et de la Communication (CReSTIC)
Université de Reims Champagne-Ardenne, Moulin de la Housse BP 1039, 51689 Reims Cedex 2, France

[2] Unité de recherche MACS, Université de Gabès, Route Médenine, 6029 Gabès, Tunisie

{dalel.jabri, kevin.guelton, noureddine.manamanni}@univ-reims.fr

**Résumé :**

Dans cet article, la stabilisation non-quadratique décentralisée des systèmes non-linéaires composés de $n$ sous systèmes descripteurs flous de type Takagi-Sugeno est abordée. Afin d'assurer la stabilité du système globale en boucle fermée et de minimiser l'effet des interconnections entre les sous systèmes, le résultat principal permet la synthèse d'un réseau de lois de commande décentralisées de type Compensations Parallèles Distribuées modifiées via un critère $H_\infty$. Les conditions de stabilité, exprimées sous la forme d'un ensemble d'Inégalités Linéaires Matricielles, sont obtenues via une fonction non quadratique de Lyapunov. Finalement, un exemple numérique illustre l'efficacité de l'approche de commande décentralisée proposée.

**Mots-clés :**

Takagi-Sugeno, Descripteurs, Interconnections, LMI, Commande décentralisée.

**Abstract:**

This paper deals with the non-quadratic decentralized stabilization of a set of $n$ Takagi-Sugeno descriptors. To ensure the stability of the whole closed-loop dynamics and to minimize interconnection effects between subsystems, the main result allows designing a network of non Parallel Distributed Compensation control laws via a $H_\infty$ criterion. Sufficient conditions, obtained from a non quadratic fuzzy Lyapunov approach, are provided in terms of Linear Matrix Inequalities. Finally, a numerical example illustrates the efficiency of the proposed decentralized control approach.

## 1 Introduction

Parmi les approches qui s'intéressent à la modélisation des systèmes non linéaires, l'approche de modélisation des systèmes dynamiques proposée par Takagi et Sugeno (T-S) a montré son efficacité. En effet, un modèle T-S est constitué d'un ensemble de modèles linéaires interconnectés par des fonctions d'appartenances non linéaires [1]. Ainsi, par le biais d'une transformation polytopique convexe, telle que le découpage en secteurs non linéaires [2], il est possible d'obtenir un représentant T-S d'un modèle non linéaire valable sur un compact de l'espace d'état. Cette particularité permet d'étendre au cas des systèmes non linéaires certains concepts relatifs aux systèmes linéaires. La stabilité de tels systèmes est souvent étudiée par la seconde approche de Lyapunov. Ainsi, de nombreuse études ont permis l'obtention des conditions sous forme d'Inégalité Linéaires Matricielles (LMI) par le biais d'une fonction candidate quadratique de Lyapunov, par exemple [2][3][4][5][6]. Malheureusement, ces approches ce sont avérées conservatives car nécessitent l'existence d'une matrice commune vérifiant un ensemble de contraintes LMIs. Ainsi, des schémas de relaxations ont été proposés [3][7]. Plus récemment, des conditions de stabilité ont été introduites sur la base de fonctions candidates non-quadratique de Lyapunov [8][9][10][11]. Celles-ci présentent l'avantage de respecter la structure d'interconnexion des modèles T-S à stabiliser et permet ainsi de réduire le conservatisme des conditions LMIs. D'autre part, la stabilisation des descripteurs T-S a fait l'objet de récentes études [11] [12][13]. Ce type de systèmes élargi la classe des systèmes non linéaires explicites classiquement étudiés dans le cadre T-S aux systèmes implicites tels que les systèmes singuliers [14] ou encore les systèmes mécanique à inerties variables [15]. Par ailleurs, avec la croissante augmentation de la complexité des systèmes et de la taille des problèmes rencontrés, de nouvelles approches de commande décentralisées pour les systèmes T-S ont été proposées [16][17][18][19]. Néanmoins, à ce jour, peu de travaux traitent du problème de la synthèse de lois de commande décentralisées stabilisant un ensemble de descripteurs T-S interconnectés. Dans [20], une première approche a été proposée mais au prix d'hypothèses réduisant l'étendue de la classe des systèmes considérés. Dans [21], des conditions de stabilité, pour une classe générique de descripteurs T-S, ont été étudiés. Toutefois, ces études ne permettent pas d'optimiser les performances de la boucle fermée.

Dans cet article, nous proposons d'améliorer ces performances en utilisant un critère $H_\infty$ afin de réduire les influences des interconnexions entre sous systèmes. Après avoir présenté la classe des systèmes étudiée ainsi que le problème de commande considéré, des conditions de stabilité seront proposées sous forme LMI. Enfin, un exemple numérique illustrera l'efficacité de l'approche proposée.

## 2 Définition du problème de commande décentralisée

Soit un ensemble $S$ de $n$ systèmes descripteurs flous de type Takagi-Sugeno $S_i$ interconnectés entre eux et décrits par :

Pour $i = 1, ..., n$ :

$$\sum_{j=1}^{l_i} v_i^j(z_i) E_i^j \dot{x}_i(t) = \sum_{k=1}^{r_i} h_i^k(z_i) \begin{pmatrix} A_i^k x_i(t) + B_i^k u_i(t) \\ + \sum_{\substack{\alpha=1 \\ \alpha \neq i}}^{n} F_{i\alpha}^k x_\alpha(t) \end{pmatrix}$$
(1)

où $x_i(t) \in \mathbb{R}^{n_i}$, $u_i(t) \in \mathbb{R}^{m_i}$ et $z_i(t) \in \mathbb{R}^{p_i}$ sont respectivement les vecteurs d'état, de commande et de prémisses associés au $i^{ème}$ modèle. $x_\alpha(t) \in \mathbb{R}^{n_\alpha}$ est le vecteur d'état du $\alpha^{ième}$ modèle avec $\alpha = 1, ..., n$ et $\alpha \neq i$. $l_i$ est le nombre de règles floues associées au membre gauche de l'équation d'état (1). Ainsi, pour $j = 1, ..., l_i$ on a $E_i^j \in \mathbb{R}^{n_i \times n_i}$ des matrices à coefficients constants, le cas échéant singulières, $v_i^j(z_i) \geq 0$ et $h_i^k(z_i) \geq 0$ les fonctions d'appartenances de classe C1 vérifiant la propriété de somme convexe $\sum_{j=1}^{l_i} v_i^j(z_i) = 1$, $\sum_{k=1}^{r_i} h_i^k(z_i) = 1$. De même, $r_i$ représente le nombre de règles floues associées au membre droit dans (1). Ainsi, pour $k = 1, ..., r_i$ on a $A_i^k \in \mathbb{R}^{n_i \times n_i}$, $B_i^k \in \mathbb{R}^{n_i \times m_i}$ et $F_{i\alpha}^k \in \mathbb{R}^{n_i \times n_\alpha}$ des matrices à coefficients constants. Notons que le terme $\sum_{\substack{\alpha=1 \\ \alpha \neq i}}^{n} F_{i\alpha}^k x_\alpha(t)$ exprime l'influence du système $\alpha$ sur la dynamique du système $i$. Enfin, chaque sous système descripteur $S_i$ est supposé défini et non impulsif [14].

Afin d'assurer la stabilisation du système $S$ dans son ensemble et de minimiser l'influence de chaque sous systèmes sur la dynamique globale du système $S_i$, l'approche non PDC (Parallel Distributed Compensation) décentralisée est proposée. L'idée est de développer un ensemble de $n$ lois de commande non PDC utilisant la même structure d'interconnexion floue que celle du modèle flou T-S pour lequel il est synthétisé et garantissant la stabilité intrinsèque de chaque sous système $S_i$ tout en tenant compte de ses interactions avec les autres sous systèmes. Ainsi, pour chaque système descripteur de type T-S $S_i$, la loi de commande décentralisée de type non PDC est décrite par :

$$u_i(t) = \left( \sum_{j=1}^{l_i} \sum_{s=1}^{r_i} v_i^j(z_i(t)) h_i^s(z_i(t)) K_i^{js} \right) \times \left( \sum_{j=1}^{l_i} \sum_{s=1}^{r_i} v_i^j(z_i(t)) h_i^s(z_i(t)) X_{i1}^{js} \right)^{-1} x_i(t) \quad (2)$$

En combinant, les équations (1) et (2), la dynamique de la boucle fermée du système globale $S$ peut être exprimée par :

Pour $i = 1,...,n$,

$$\sum_{j=1}^{l_i} v_i^j E_i^j \dot{x}_i = \sum_{j=1}^{l_i} \sum_{k=1}^{r_i} \sum_{s=1}^{r_i} v_i^j h_i^k h_i^s G_i^{jks}(v_i^j, h_i^s) \quad (3)$$

avec

$$G_i^{jks}(v_i^j, h_i^s) = \left( A_i^k + B_i^k K_i^{js} \left( \sum_{j=1}^{l_i} \sum_{s=1}^{r_i} v_i^j h_i^s X_{i1}^{js} \right)^{-1} \right) x_i + \sum_{\substack{\alpha=1 \\ \alpha \neq i}}^{n} F_{i\alpha}^k x_\alpha$$

Dans la suite de cet article, afin d'alléger les écritures mathématiques, les notations suivantes seront utilisées.

**Notations :**

$$E_i^v = \sum_{j=1}^{l_i} v_i^j(z_i) E_i^j, \quad Y_i^{hh} = \sum_{j=1}^{l_i} \sum_{k=1}^{l_i} h_i^j(z_i) h_i^k(z_i) Y_i^{jk}$$

$$T_i^{vhh} = \sum_{j=1}^{l_i} \sum_{s=1}^{r_i} \sum_{k=1}^{r_i} v_i^j(z_i) h_i^s(z_i) h_i^k(z_i) T_i^{jsk}, \quad \text{ainsi de suite...}$$

Une étoile (*) indique une quantité transposée dans les écritures matricielles. De plus, le temps $t$ sera omit lorsqu'il n'y aura pas d'ambiguïté.

# 3 Synthèse H∞ de lois de commande décentralisées

L'objectif est maintenant de fournir une méthodologie de synthèse de lois de commande décentralisées (2) stabilisant les descripteurs interconnectés décrits par (1) tout en réduisant l'effet des interconnections entre sous systèmes. Pour ce faire, on considère le critère $H_\infty$ suivant :

$$\int_{t_0}^{t_f} x_i^T x_i \, dt < \rho_i^2 \int_{t_0}^{t_f} \varphi_{i\alpha}^T \varphi_{i\alpha} \, dt \quad (4)$$

avec $\varphi_{i\alpha}(z_i) = \sum_{k=1}^{r_i} \sum_{\substack{\alpha=1 \\ \alpha \neq i}}^{n} h_i^k F_{i\alpha}^k x_\alpha$ le vecteur traduisant l'influence du $\alpha^{ème}$ descripteur sur le descripteur $i$ et $\rho_i$ les taux de performance $H\infty$.

Le résultat principal de cet article est résumé par le théorème suivant.

**Théorème 1** :
Soit, pour tout $i = 1,...,n$, $j = 1,...,l_i$ et $s = 1,...,r_i$, $\dot{h}_i^s(z(t)) \geq \varpi_i^s$ et $\dot{v}_i^j(z(t)) \geq \lambda_i^j$. L'ensemble $S$ des $n$ descripteurs T-S interconnectés $S_i$ décrits par (1) est globalement asymptotiquement stable en boucle fermée via le réseau de $n$ lois de commande non PDC décentralisées défini par (1) au regard du critère $H_\infty$ décrit en (4) s'il existe les matrices $X_{i1}^{js} = \left(X_{i1}^{js}\right)^T > 0$, $X_{i3}^{ks}$, $X_{i4}^{ks}$, $K_i^{js}$ et les scalaires $\rho_i$ pour toutes les combinaisons $i = 1,...,n$, $\alpha = 1,...,n$, $\alpha \neq i$, $j = 1,...,l_i$, $k = 1,...,r_i$, $s = 1,...,r_i$, vérifiant les conditions LMIs données par :

Minimiser $\rho_i$ tel que:

$$\Gamma_{i\alpha}^{jkk} < 0 \quad (5)$$

$$\frac{1}{r_i - 1}\Gamma_{i\alpha}^{jkk} + \frac{1}{2}\left(\Gamma_{i\alpha}^{jsk} + \Gamma_{i\alpha}^{jks}\right) < 0 \qquad (6)$$

avec

$$\Gamma_{i\alpha}^{jks} = \begin{bmatrix} \Gamma_{i(1,1)}^{jks} & & & (*) \\ \Gamma_{i(2,1)}^{jks} & \Gamma_{i(2,2)}^{jks} & & \\ 0 & (n-1)\left(F_{i\alpha}^{k}\right)^T & \Gamma_{i\alpha(3,3)}^{k} & \\ \left(X_{i1}^{js}\right)^T & 0 & 0 & -I \end{bmatrix},$$

$$\Gamma_{i(1,1)}^{jks} = X_{i3}^{ks} + \left(X_{i3}^{ks}\right)^T - \sum_{s=1}^{r_i}\varpi_i^s X_{i1}^{js} - \sum_{s=1}^{l_i}\lambda_i^j X_{i1}^{js},$$

$$\Gamma_{i(2,1)}^{jks} = A_i^k X_{i1}^{js} + B_i^k K_i^{js} - E_i^j X_{i3}^{ks} + \left(X_{i4}^{ks}\right)^T,$$

$$\Gamma_{i(2,2)}^{jks} = -E_i^j X_{i4}^{ks} - \left(E_i^j\right)^T\left(X_{i4}^{ks}\right)^T$$

et $\Gamma_{i\alpha(3,3)}^{k} = -\rho_i^2(n-1)(2n-3)\left(F_{i\alpha}^{k}\right)^T F_{i\alpha}^{k}$.

***Preuve :***

On pose, pour $i=1,\ldots,n$, $\tilde{x}_i = \begin{bmatrix} x_i & \dot{x}_i \end{bmatrix}^T$ les vecteurs d'états étendus. Le critère $H\infty$ (4) peut être réécrit sous la forme :

$$\int_{t_0}^{t_f}\tilde{x}_i^T\tilde{Q}_i\tilde{x}_i dt < \rho_i^2\int_{t_0}^{t_f}\sum_{\substack{\alpha=1 \\ \alpha\neq i}}^{n}\left(\begin{array}{l}(n-1)\left(\tilde{x}_\alpha^T\left(\tilde{F}_{i\alpha}^h\right)^T \tilde{F}_{i\alpha}^h\tilde{x}_\alpha\right) \\ + \sum_{\substack{\beta=1 \\ \beta\neq i \\ \beta\neq\alpha}}^{n}\left(\tilde{x}_\beta^T\left(\tilde{F}_{i\beta}^h\right)^T \tilde{F}_{i\beta}^h\tilde{x}_\beta\right)\end{array}\right)dt$$

(7)

avec $\tilde{Q}_i = \begin{bmatrix} I & 0 \\ 0 & 0 \end{bmatrix}$.

De même, (3) peut être réécrite, avec les notations définies précédemment, sous la forme étendue donnée par :

Pour $i=1,\ldots,n$, $\tilde{E}\dot{\tilde{x}}_i = \sum_{\substack{\alpha=1 \\ \alpha\neq i}}^{n}\left(\tilde{A}_i^{vhh}\tilde{x}_i + \tilde{F}_{i\alpha}^h\tilde{x}_\alpha\right)$ (8)

avec $\tilde{E} = \begin{bmatrix} I & 0 \\ 0 & 0 \end{bmatrix}$, $\tilde{F}_{i\alpha}^h = \begin{bmatrix} 0 & 0 \\ F_{i\alpha}^h & 0 \end{bmatrix}$.

et $\tilde{A}_i^{vhh} = \frac{1}{n-1}\begin{bmatrix} 0 & I \\ A_i^h + B_i^h K_i^{vh}\left(X_{i1}^{vh}\right)^{-1} & -E_i^v \end{bmatrix}$.

Soit la fonction candidate non quadratique de Lyapunov multiple donnée par :

$$V(t) = \sum_{i=1}^{n}V_i\left(x_i(t)\right) \geq 0 \qquad (9)$$

avec $V_i\left(x_i(t)\right) = \tilde{x}_i^T(t)\tilde{E}\left(\tilde{X}_i^{vhh}\right)^{-1}\tilde{x}_i(t)$ et la condition de symétrie $\tilde{E}X_i^{-1} = X_i^{-T}\tilde{E} \geq 0$ conduisant au conditionnement $\tilde{X}_i^{vhh} = \begin{bmatrix} X_{i1}^{vh} & 0 \\ X_{i3}^{hh} & X_{i4}^{hh} \end{bmatrix}$ et $X_{i1}^{vh} = \left(X_{i1}^{vh}\right)^T > 0$.

Le système (8) est stable en boucle fermée selon le critère $H\infty$, si :

$$\sum_{i=1}^{n}\left(\dot{V}_i(t) + \tilde{x}_i^T\tilde{Q}_i\tilde{x}_i - \rho_i^2\varphi_i(z_i)^T\varphi_i(z_i)\right) < 0 \quad (10)$$

C'est-à-dire si :

$$\sum_{i=1}^{n}\left(\begin{array}{l}\tilde{x}_i^T\left(\begin{array}{l}(n-1)\left(\begin{array}{l}\left(\tilde{A}_i^{vhh}\right)^T\left(\tilde{X}_i^{vhh}\right)^{-1} \\ +\left(\tilde{X}_i^{vhh}\right)^{-T}\tilde{A}_i^{vhh}\end{array}\right) \\ +\tilde{Q}_i + \tilde{E}\overline{\left(\tilde{X}_i^{vhh}\right)^{-1}}\end{array}\right)\tilde{x}_i \\ +\sum_{\substack{\alpha=1 \\ \alpha\neq i}}^{n}\left(\begin{array}{l}\tilde{x}_\alpha^T\left(\tilde{F}_{i\alpha}^h\right)^T\left(\tilde{X}_i^{vhh}\right)^{-1}\tilde{x}_i \\ +\tilde{x}_i^T\left(\tilde{X}_i^{vhh}\right)^{-T}\tilde{F}_{i\alpha}^h\tilde{x}_\alpha\end{array}\right) \\ -\rho_i^2\left(\begin{array}{l}(n-1)\sum_{\substack{\alpha=1 \\ \alpha\neq i}}^{n}\tilde{x}_\alpha^T\left(\tilde{F}_{i\alpha}^h\right)^T\tilde{F}_{i\alpha}^h\tilde{x}_\alpha \\ +\sum_{\substack{\alpha=1 \\ \alpha\neq i}}^{n}\sum_{\substack{\beta=1 \\ \beta\neq i \\ \beta\neq\alpha}}^{n}\tilde{x}_\beta^T\left(\tilde{F}_{i\beta}^h\right)^T\tilde{F}_{i\beta}^h\tilde{x}_\beta\end{array}\right)\end{array}\right) < 0 \quad (11)$$

Notons que $\sum_{\substack{\alpha=1 \\ \alpha\neq i}}^{n}\left(\Psi_{i\alpha}+\sum_{\substack{\beta=1 \\ \beta\neq i \\ \beta\neq \alpha}}^{n}\Psi_{i\beta}\right)=(n-1)\sum_{\substack{\alpha=1 \\ \alpha\neq i}}^{n}\Psi_{i\alpha}$, l'inégalité (11) devient:

$$\sum_{i=1}^{n}\sum_{\substack{\alpha=1 \\ \alpha\neq i}}^{n}\begin{pmatrix}\tilde{x}_i^T\left(\begin{pmatrix}\left(\tilde{A}_i^{vhh}\right)^T\left(\tilde{X}_i^{vhh}\right)^{-1}+\left(\tilde{X}_i^{vhh}\right)^{-T}\tilde{A}_i^{vhh}\\+\dfrac{1}{(n-1)}\left(\tilde{Q}_i+\tilde{E}\overline{\left(\tilde{X}_i^{vhh}\right)^{-1}}\right)\end{pmatrix}\tilde{x}_i\\+\tilde{x}_\alpha^T\left(\tilde{F}_{i\alpha}^h\right)^T\left(\tilde{X}_i^{vhh}\right)^{-1}\tilde{x}_i+\tilde{x}_i^T\left(\tilde{X}_i^{vhh}\right)^{-T}\tilde{F}_{i\alpha}^h\tilde{x}_\alpha\\-\rho_i^2(2n-3)\tilde{x}_\alpha^T\left(\tilde{F}_{i\alpha}^h\right)^T\tilde{F}_{i\alpha}^h\tilde{x}_\alpha\end{pmatrix}<0 \quad (12)$$

Quel l'on peut réécrire sous la forme :

$$\sum_{i=1}^{n}\sum_{\substack{\alpha=1 \\ \alpha\neq i}}^{n}\begin{bmatrix}\tilde{x}_i\\\tilde{x}_\alpha\end{bmatrix}^T\begin{bmatrix}\Omega_{i(1,1)}^{vhh} & (*)\\\left(\tilde{F}_{i\alpha}^h\right)^T\left(\tilde{X}_i^{vhh}\right)^{-1} & \Omega_{i(2,2)}^h\end{bmatrix}\begin{bmatrix}\tilde{x}_i\\\tilde{x}_\alpha\end{bmatrix}<0 \quad (13)$$

avec $\Omega_{i(1,1)}^{vhh}=\begin{pmatrix}\left(\tilde{A}_i^{vhh}\right)^T\left(\tilde{X}_i^{vhh}\right)^{-1}+\left(\tilde{X}_i^{vhh}\right)^{-T}\tilde{A}_i^{vhh}\\+\dfrac{1}{(n-1)}\left(\tilde{Q}_i+\tilde{E}\overline{\left(\tilde{X}_i^{vhh}\right)^{-1}}\right)\end{pmatrix}$

et $\Omega_{i(2,2)}^h=-\rho_i^2(2n-3)\left(\tilde{F}_{i\alpha}^h\right)^T\tilde{F}_{i\alpha}^h$.

Donc, (13) est vérifiée si, pour tout $i=1,...,n$, $\alpha=1,...,n$, $\alpha\neq i$ :

$$\begin{bmatrix}\Omega_{i(1,1)}^{vhh} & (*)\\\left(\tilde{F}_{i\alpha}^h\right)^T\left(\tilde{X}_i^{vhh}\right)^{-1} & \Omega_{i(2,2)}^h\end{bmatrix}<0 \quad (14)$$

En multipliant (14) à gauche et à droite respectivement par $\begin{bmatrix}\left(\tilde{X}_i^{vhh}\right)^T & 0\\0 & I\end{bmatrix}$, on obtient, pour tout $i=1,...,n$, $\alpha=1,...,n$, $\alpha\neq i$ :

$$\begin{bmatrix}\Lambda_{i(1,1)}^{vhh} & (*)\\\left(\tilde{F}_{i\alpha}^h\right)^T & \Omega_{i(2,2)}^h\end{bmatrix}<0 \quad (15)$$

avec

$$\Lambda_{i(1,1)}^{vhh}=\begin{pmatrix}\left(\tilde{X}_i^{vhh}\right)^T\left(\tilde{A}_i^{vhh}\right)^T+\tilde{A}_i^{vhh}\tilde{X}_i^{vhh}\\+\dfrac{1}{(n-1)}\left(\tilde{X}_i^{vhh}\right)^T\left(\tilde{Q}_i+\tilde{E}\overline{\left(\tilde{X}_i^{vhh}\right)^{-1}}\right)\tilde{X}_i^{vhh}\end{pmatrix}$$

Notons que $\overline{\left(\tilde{X}_i^{vhh}\right)^{-1}}=-\left(\tilde{X}_i^{vhh}\right)^{-1}\dot{\tilde{X}}_i^{vhh}\left(\tilde{X}_i^{vhh}\right)^{-1}$ (voir par exemple [21]), on peut écrire :

$$\Lambda_{i(1,1)}^{vhh}=\begin{pmatrix}\left(\tilde{X}_i^{vhh}\right)^T\left(\tilde{A}_i^{vhh}\right)^T+\tilde{A}_i^{vhh}\tilde{X}_i^{vhh}\\+\dfrac{1}{(n-1)}\left(\left(\tilde{X}_i^{vhh}\right)^T\tilde{Q}_i\tilde{X}_i^{vhh}-\tilde{E}\overline{\left(\tilde{X}_i^{vhh}\right)}\right)\end{pmatrix} \quad (16)$$

Le terme $-\overline{\left(\dot{X}_{1i}^{vh}\right)}$ peut être réécrit en utilisant les propriétés des sommes convexes tel que:

$$\overline{\left(\dot{X}_{1i}^{vh}\right)}=\sum_{j=1}^{l_i}\sum_{s=1}^{r_i}v_i^j h_i^s\left(\sum_{s=1}^{r_i}\dot{h}_i^s X_{i1}^{js}+\sum_{s=1}^{l_i}\dot{v}_i^j X_{i1}^{js}\right) \quad (17)$$

Soient $\lambda_i^j$ et $\varpi_i^s$, pour tout $j=1,...,l_i$ et $s=1,...,r_i$, respectivement les bornes inférieurs des $\dot{v}_i^j(z)$ et $\dot{h}_i^s(z)$. $-\overline{\left(\dot{X}_{1i}^{vh}\right)}$ peut être majoré par:

$$-\overline{\left(\dot{X}_{1i}^{vh}\right)}\leq -\Phi_{hv} \quad (18)$$

avec $\Phi_{hv}=\sum_{j=1}^{l_i}\sum_{s=1}^{r_i}v_i^j h_i^s\left(\sum_{s=1}^{r_i}\varpi_i^s X_{i1}^{js}+\sum_{s=1}^{l_i}\lambda_i^j X_{i1}^{js}\right)$

L'inégalité (15), majorée par (18) peut donc être réécrite sous sa forme étendue avec les matrices étendues définies aux sein des expressions (7), (8), (9). Dès lors, en appliquant le complément de Schur, on obtient, pour tout $i=1,...,n$, $\alpha=1,...,n$, $\alpha \neq i$ :

$$\Gamma_{i\alpha}^{vhh} = \begin{bmatrix} \Gamma_{i(1,1)}^{vhh} & & & (*) \\ \Gamma_{i(2,1)}^{vhh} & \Gamma_{i(2,2)}^{vhh} & & \\ 0 & (n-1)\left(\tilde{F}_{i\alpha}^h\right)^T & \Gamma_{i\alpha(3,3)}^h & \\ \left(X_{i1}^{vh}\right)^T & 0 & 0 & -I \end{bmatrix} < 0$$
(19)

avec $\Gamma_{i(1,1)}^{vhh} = X_{i3}^{hh} + \left(X_{i3}^{hh}\right)^T - \Phi_{hv}$,

$\Gamma_{i(2,1)}^{vhh} = A_i^h X_{i1}^{vh} + B_i^h K_i^v - E_i^v X_{i3}^{hh} + \left(X_{i4}^{hh}\right)^T$,

$\Gamma_{i(2,2)}^{vhh} = -E_i^v X_{i4}^{hh} - \left(E_i^v\right)^T \left(X_{i4}^{hh}\right)^T$

et $\Gamma_{i\alpha(3,3)}^h = -\rho_i^2 (n-1)(2n-3)\left(\tilde{F}_{i\alpha}^h\right)^T \tilde{F}_{i\alpha}^h$.

Enfin, en appliquant le schéma de relaxation proposée dans [3], on obtient les conditions LMI (5) et (6) proposées au théorème 1. ∎

*Remarque* : Les conditions non quadratiques proposées au théorème 1 nécessitent la connaissance à priori des bornes inférieures des dérivées des fonctions d'appartenances. L'obtention de ces bornes n'est pas aisée dans la pratique. De ce fait, dans le cadre de la commande des systèmes descripteurs, un compromis peut-être proposé en supposant certaines variables de décisions communes à chaque LMI [22]. Dans le cadre de la commande décentralisée, le résultat est proposé au corollaire suivant.

**Corollaire 1**
L'ensemble $S$ des $n$ descripteurs T-S interconnectés $S_i$ décrits par (1) est globalement asymptotiquement stable en boucle fermée via le réseau de $n$ lois de commande non PDC décentralisées défini par (1), au regard du critère $H_\infty$ décrit en (4), s'il existe les matrices $X_{i1} = \left(X_{i1}\right)^T > 0$, $X_{i3}^{ks}$, $X_{i4}^{ks}$, $K_i^{js}$ et les scalaires $\rho_i$ pour toutes les combinaisons $i=1,...,n$, $\alpha=1,...,n$, $\alpha \neq i$, $j=1,...,l_i$, $k=1,...,r_i$, $s=1,...,r_i$, vérifiant les conditions LMIs données par :

Minimiser $\rho_i$ tel que:

$$T_{i\alpha}^{jkk} < 0 \qquad (20)$$

$$\frac{1}{r_i - 1} T_{i\alpha}^{jkk} + \frac{1}{2}\left(T_{i\alpha}^{jsk} + T_{i\alpha}^{jks}\right) < 0 \qquad (21)$$

avec

$$T_{i\alpha}^{jks} = \begin{bmatrix} T_{i(1,1)}^{ks} & & & (*) \\ T_{i(2,1)}^{jks} & T_{i(2,2)}^{jks} & & \\ 0 & (n-1)\left(F_{i\alpha}^k\right)^T & T_{i\alpha(3,3)}^k & \\ X_{i1}^T & 0 & 0 & -I \end{bmatrix},$$

$T_{i(1,1)}^{ks} = X_{i3}^{ks} + \left(X_{i3}^{ks}\right)^T$,

$T_{i(2,1)}^{jks} = A_i^k X_{i1} + B_i^k K_i^{js} - E_i^j X_{i3}^{ks} + \left(X_{i4}^{ks}\right)^T$,

$T_{i(2,2)}^{jks} = -E_i^j X_{i4}^{ks} - \left(E_i^j\right)^T \left(X_{i4}^{ks}\right)^T$

et $T_{i\alpha(3,3)}^k = -\rho_i^2 (n-1)(2n-3)\left(F_{i\alpha}^k\right)^T F_{i\alpha}^k$.

**Preuve :**
Triviale en suivant le chemin donné par la démonstration du théorème 1 avec $X_{i1}$ commune pour tous $j=1,...,l_i$, $s=1,...,r_i$. ∎

## 4 Exemple numérique

Soit deux descripteurs $S_1$ et $S_2$ interconnectés de type T-S décrits par :

$$S_i \bigg|_{\substack{i=1,2 \\ \alpha=1,2 \\ \alpha \neq i}} \begin{vmatrix} \sum_{j=1}^{2} v_i^j\left(x_i(t)\right) E_i^j \dot{x}_i(t) \\ = \sum_{k=1}^{2} h_i^k\left(x_i(t)\right)\left(A_i^k x_i(t) + B_i^k u_i(t) + F_{i\alpha}^k x_\alpha(t)\right) \end{vmatrix}$$
(22)

avec $E_1^1 = \begin{bmatrix} 1 & 0 \\ -1 & 1 \end{bmatrix}$, $E_1^2 = \begin{bmatrix} 1 & 0.5 \\ -1 & 1 \end{bmatrix}$,

$A_1^1 = \begin{bmatrix} a & 1 \\ 0.839 & -0.73 \end{bmatrix}$, $A_1^2 = \begin{bmatrix} -0.73 & 1 \\ 0.839 & -0.73 \end{bmatrix}$,

$B_1^1 = \begin{bmatrix} 0.47 \\ 1.263 \end{bmatrix}$, $B_1^2 = \begin{bmatrix} b \\ 1.263 \end{bmatrix}$, $F_{12}^1 = \begin{bmatrix} 0 & 0 \\ 0.1 & 0 \end{bmatrix}$,

$F_{12}^2 = \begin{bmatrix} 0 & 0 \\ 0.2 & 0 \end{bmatrix}$, $v_1^1(x_1(t)) = \left( \frac{1 - \cos(x_{11})}{2} \right)$,

$v_1^2(x_1(t)) = 1 - v_1^1$, $h_1^1(x_1(t)) = \sin^2(x_{11}(t))$,

$h_1^2(x_1(t)) = 1 - h_1^1$ et avec $E_2^1 = \begin{bmatrix} 1 & 0.2 \\ 0 & 1 \end{bmatrix}$,

$E_1^2 = \begin{bmatrix} 1 & 0 \\ 0 & 1 \end{bmatrix}$, $A_2^1 = \begin{bmatrix} -1 & 1 \\ 0.839 & -0.931 \end{bmatrix}$,

$A_2^2 = \begin{bmatrix} -1 & 1 \\ 0 & 0.931 \end{bmatrix}$, $B_2^1 = \begin{bmatrix} 0.47 \\ 0.4 \end{bmatrix}$,

$B_2^2 = \begin{bmatrix} 0.47 \\ 0.8 \end{bmatrix}$, $F_{21}^1 = \begin{bmatrix} 0 & 0 \\ 0.3 & 0 \end{bmatrix}$, $F_{21}^2 = \begin{bmatrix} 0 & 0 \\ 0.5 & 0 \end{bmatrix}$,

$v_2^1(x_2(t)) = \cos^2(x_{21}(t))$, $v_2^2(x_2(t)) = 1 - v_2^1$,

$h_2^1(x_2(t)) = \sin^2(x_{21}(t))$, $h_2^2(x_2(t)) = 1 - h_2^1$.

La figure 1 présente une comparaison des domaines de faisabilité obtenus via l'approche non quadratique notée (théorème 1) et l'approche quadratique relâché notée (corollaire 1).

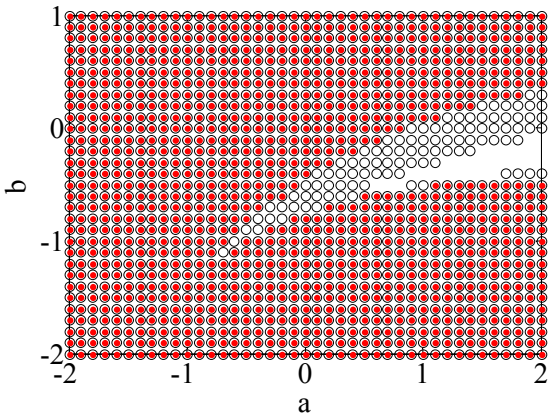

Figure 1 – domaines de faisabilité
(•) Corollaire 1, (o) Théorème 1.

La figure 2 présente une comparaison des domaines de faisabilités obtenus via les conditions LMI du théorème 1 et celles proposées dans [21]. Notons que, dans cette dernière, aucun critère de performance n'a été employé pour la synthèse des correcteurs. De plus, on souligne qu'il n'existe pas de lemme d'inclusion entre ces deux approches puisque leurs conditions LMI ne sont pas basées sur les mêmes majorations.

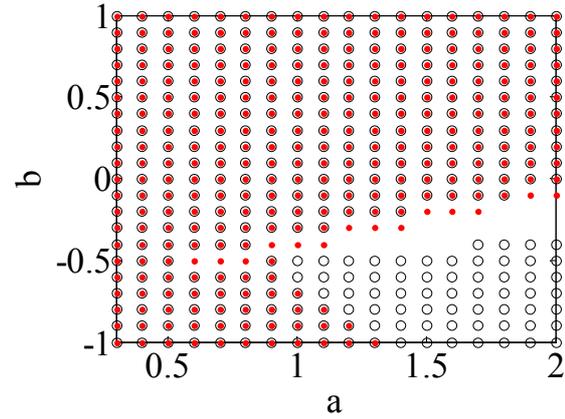

Figure 2 – Domaines de faisabilité (•) [21], (o)Théorème 1

Afin d'illustrer les performances de la synthèse $H_\infty$ proposée, la figure 3 présente la simulation de (22) en boucle fermée. Les gains ont été obtenus via le théorème 1 à l'aide de la boite à outils Matlab LMI Toolbox avec les paramètres $a = 0$, $b = -0.5$, $\varpi_i^s = -2$ et $\lambda_i^j = -2$ pour tous $i = 1,2$, $s = 1,2$ et $j = 1,2$. De même, les performances $H_\infty$ obtenues sont données par $\rho_1 = 3.575$ et $\rho_2 = 3.559$.

## 5  Conclusion

Dans cette étude, une méthodologie de synthèse de contrôleurs flous décentralisés est proposée via l'utilisation d'un critère $H_\infty$ pour la classe des descripteurs interconnectés de type T-S. Une fonction candidate non quadratique de Lyapunov multiple ainsi qu'un réseau de $n$ lois de commande non-PDC ont été employés afin d'aboutir à des conditions LMIs. Enfin, un exemple numérique a permis d'illustrer l'efficacité de l'approche proposée.

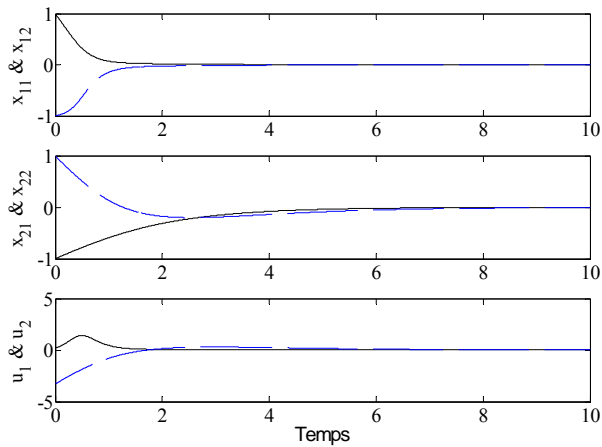

Figure 2 – Simulation en boucle fermée, (x11, x21, u1) solide, (x12, x22, u2) pointillé.

## Références